# Classifying Images with CoLaNET Spiking Neural Network – the MNIST Example

**Mikhail Kiselev**
Kaspersky
Cheboxary, Russia
`mikhail.kiselev@kaspersky.com`

September 10, 2024

## Abstract

In the present paper, it is shown how the columnar/layered CoLaNET spiking neural network (SNN) architecture can be used in supervised learning image classification tasks. Image pixel brightness is coded by the spike count during image presentation period. Image class label is indicated by activity of special SNN input nodes (one node per class). The CoLaNET classification accuracy is evaluated on the MNIST benchmark. It is demonstrated that CoLaNET is almost as accurate as the most advanced machine learning algorithms (not using convolutional approach).

## 1    Introduction

In my recent work [1] I described a novel spiking neural network (SNN) architecture called CoLaNET (Columnar Layered NETwork). The distinctive feature of this architecture is a combination of prototypical network structures corresponding to different classes and significantly distinctive instances of one class (=columns) and functionally differing populations of neurons inside columns (=layers). The other distinctive feature is a novel combination of anti-Hebbian and dopamine-modulated plasticity. The plasticity rules are local and do not use the backpropagation principle.

As an example, I considered in [1] an application of CoLaNET to a binary classification problem where the network should learn to distinct the external world states close to the target states in a reinforcement learning task (the ping pong ATARI game). In the present paper, we consider how CoLaNET can be used in an image classification task with multiple classes. The MNIST dataset [2] is selected for this purpose as a well known benchmark representing a non-trivial classification problem with numerous reported results to compare with.

## 2    Specifics of CoLaNET Application to Image Classification

CoLaNET requires that all data are represented in the spiking form. For image classification, it means that each image is presented during certain time period, every image pixel corresponds to an SNN input node and, during this period, every input node emits number of spikes proportional to the pixel brightness. In our case, the image presentation period equals to 10 SNN iteration (we take one iteration as 1 msec). Therefore, for pixel with the brightness $b$, the emitted spike count will be $[10b/255]$.

Since all images in MNIST are independent, in order to avoid undesired influence of the previous image, the consecutive image presentations should be separated by a period when input nodes encoding pixel brightness do not emit spikes (see the discussion in [1]). This period of silence is also 10 msec.

The class labels are also encoded by spikes. CoLaNET has special input nodes corresponding to image classes – one node per class. While an image belonging to a certain class is presented, the respective input node emits spikes with highest possible frequency. This activity covers silence periods as well.

This spiking encoding method is not specific for MNIST. It is used for any collection of images where neighboring images are independent.

## 3 The Test Classification Task – MNIST.

Many image classification algorithms are tested on MNIST [2]. It is a collection of 70000 28x28 pixel monochrome images of handwritten digits, 60000 of which are used for training a classifier while the rest 10000 are used for testing. The method for conversion of MNIST data to the spiking form was described in the previous section. As it was said, each case classified should be represented by activity of input nodes (there are 28 * 28 = 768 of them) during a 10 msec period followed by 10 msec period of silence. Thus, the whole MNIST dataset is represented by 70000 * (10 + 10) = 1400000 msec of input node activity record. The first 1200000 msec were used for training; after that synaptic plasticity in CoLaNET was switched off and it worked in the inference regime. During this testing period, classification accuracy was measured (in terms of absolute error).

The CoLaNET architecture has a very few hyperparameters (see Table 1). In the case of MNIST, they are even fewer than in the example from [1], since there are no training examples in MNIST which do not belong to any class as was in [1]. In this case, the anti-Hebbian and dopamine plasticity mechanisms should be balanced and, therefore, $d_D = d_H$. In order to find the hyperparameter optimum values I used an optimization procedure based on genetic algorithm. To diminish the probability of accidental bad or good result, I averaged the criterion (the absolute error value) for 4 tests with the same SNN parameters. Since the WTA mechanism introduces non-determinism in the simulation process, the synaptic weights learnt in these tests were different. I used the typical genetic algorithm settings, used by me in my previous studies. The population size was 100, mutation probability per individual equaled to 0.5, elitism level was 0.1. Genetic algorithm terminated after 3 consecutive populations without optimization progress.

The parameters varied in the optimization procedure, the ranges of their variation and their optimum values found are presented in Table 1.

**Table 1. The CoLaNET hyperparameters optimized for the MNIST dataset.**

| Parameter optimized | Value range | Optimum value |
|---|---|---|
| Learning rate $d_D = d_H$ | 0.0004 – 0.1 (everywhere in this table the value of the threshold membrane potential is taken equal to 1) | 0.005 |
| Maximum input weight $w_{max}$ | 0.004 – 0.4 | 0.1 |
| Maximum input weight $w_{min}$ | -0.0004 – -0.4 | -0.082 |
| Number of microcolumns in one column | 1 – 30 | 15 |

The result was reached at the 5th generation. The winning network showed accuracy equal to 94.35% ± 0.16%. The configuration of the winning SNN written in ArNI-X language [3] can be found in Appendix A.

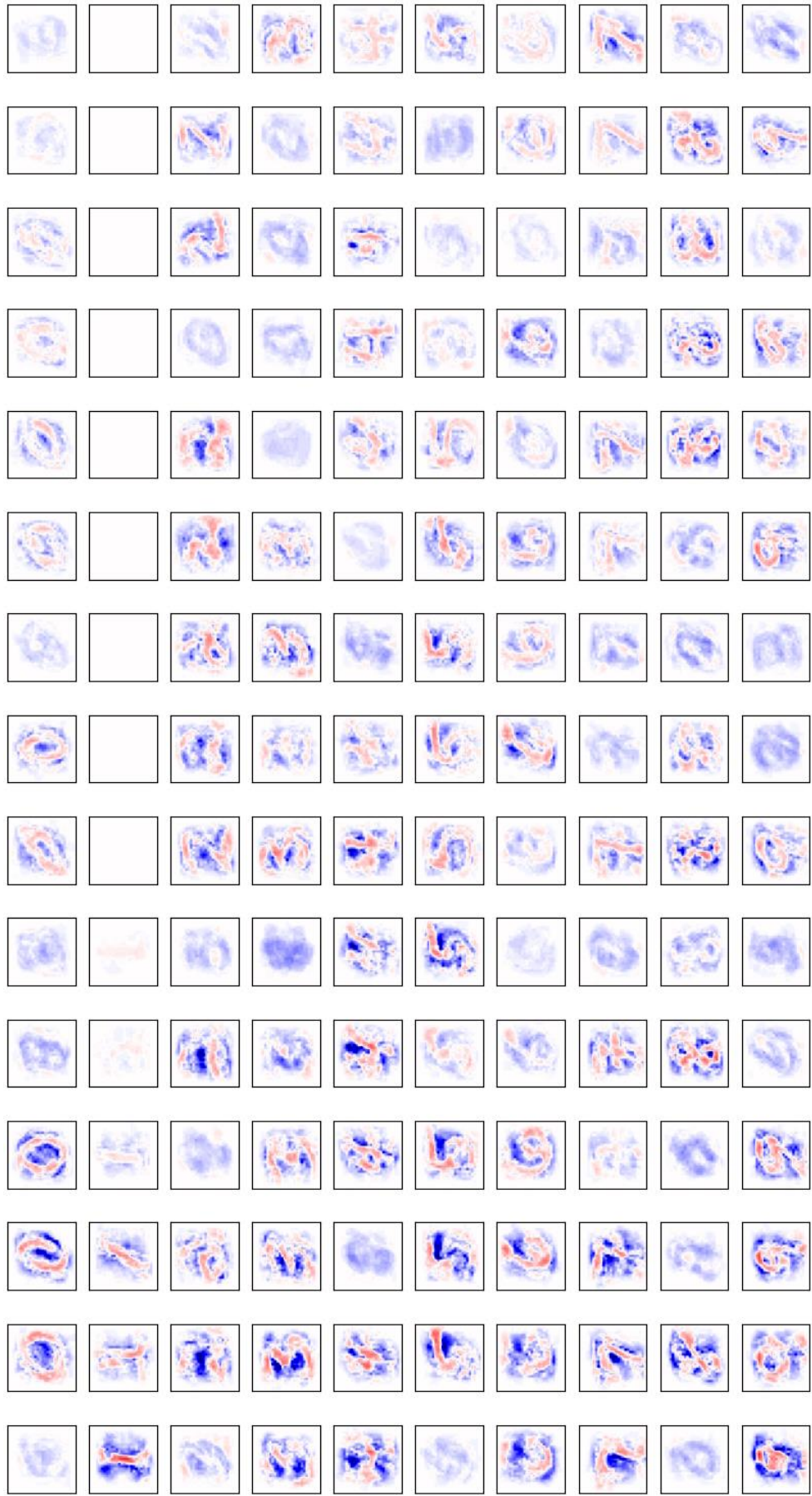

Figure 1. The learnt weights of neurons in the best SNN at the simulation end (the picture is rotated counter-clockwise). Red color corresponds to positive weights, blue – to negative weights.

The weights learnt in this task are depicted on Figure 1. Each square corresponds to one learning neuron. The picture columns correspond to CoLaNET columns (target classes, digits), the rows – to microcolumns (significantly different writings of one digit). We see that some digits have many variants of writing (like 9), while others show less variability (like 1). It is not a problem – in the case of little variability, some microcolumns simply are not used – they are always suppressed by their more lucky neighbors. They only waste the computational resources (but not so much – silent neurons consume little processor time) but do not degrade SNN accuracy.

The obtained result seems not very impressive comparatively to the SOTA result exceeding 99%. However, we should take into account that this version of CoLaNET does not use spatial structure of the images as do neural networks showing 99% accuracy. The network considered here does not include convolution layers which produce informative features from image spatial structure, - it treats all pixel values as independent predictors. Convolutional layers can be inserted in CoLaNET but it is a subject of a separate study. In order to carry out more fair comparison with other classification methods not using convolutions, I represented MNIST data in the form of rectangular table, one pixel corresponding to one column, and applied classic machine learning algorithms – random forest and multilayer perceptron (from the Sklearn package [4] – see the respective script in Appendix B). Creating these classifiers, I specified their parameters so that to make their complexity approximately equal to the complexity of the winning CoLaNET (150 * 28 * 28 weights). Perceptron showed accuracy 96.87%, random forest (often considered as the most accurate classic ML algoritms for the wide range of applications) – 95.58%. Thus, performance of CoLaNET is quite close to the best modern ML algorithms.

# 4     Result Reproducibility

To help reproducing the results reported, I created an archive file which can be downloaded from `https://disk.yandex.ru/d/jpKp7df--PsiTQ`. It contains MNIST benchmark data in the files `MNIST.bin` and `MNIST.target`. `MNIST.bin` contains MNIST images in the binary form (768 bytes per image – one after one). `MNIST.target` contains class labels – 1 label per line. The best network configuration written in ArNI-X language is contained in the file `5015.nnc`. The ArNI-X SNN simulator can be obtained from me by request. Fig. 1 is drawn by the python script `specific.py` from the simulation monitoring data `monitoring.5015.csv`. The comparison with the traditional ML algorithms is made in the script `PictureClassifierBaseline.py`.

# 5     Conclusion.

In this paper, I describe an application of CoLaNET network to image classification problems taking as an example the MNIST dataset. The accuracy demonstrated by CoLaNET was found to be close to the best modern ML algorithms not using convolutions.

Combination of CoLaNET with spiking convolutional structures is one of possible directions of further research works. Also, CoLaNET will be tested on diverse classification problems from various application fields.

Besides that, the practical implementation of CoLaNET on the AltAI neuroprocessor [5] is also planned in the future. It will be made on the new software platform – Kaspersky Neuromorhpic Platform (KNP).

# Acknowledgements.

The present work is a part of the research project on applications of SNN and AltAI neuroprocessor to practical problems carried out by Kaspersky Neuromorphic AI Team.

The computations reported in the paper were performed on a GPU-equipped computer belonging Kaspersky using the SNN simulator package ArNI-X. At present, these results are being reproduced on the KNP platform.

## References.

# Appendix A. The Configuration of the Best SNN Solved the MNIST Problem

```xml
<?xml version="1.0" encoding="utf - 8"?>
<SNN>
  <Global>0</Global>
  <Global>0.023817</Global>
  <RECEPTORS name="R" n="784">
    <Implementation lib="fromFile">
      <args type="image">
        <source>MNIST.bin</source>
        <Special>
        <width>28</width>
        <height>28</height>
        <offset>0</offset>
        <ntact_per_image>20</ntact_per_image>
        <image_presentation_time>10</image_presentation_time>
        <maxfrequency>1.</maxfrequency>
        </Special>
      </args>
    </Implementation>
  </RECEPTORS>
  <RECEPTORS name="Target" n="10">
    <Implementation lib="StateClassifier">
      <args>
        <target_file>MNIST.target</target_file>
        <spike_period>1</spike_period>
        <state_duration>20</state_duration>
        <learning_time>1200000</learning_time>
      <no_class>-</no_class>
      <prediction_file>restmp.csv</prediction_file>
      </args>
    </Implementation>
  </RECEPTORS>
<NETWORK ncopies="15">
    <Sections>
        <Section name="L">
          <props>
        <n>150</n>
        <Structure type="O" dimension="10"></Structure>
        <chartime>3</chartime>
        <weight_inc>-0.042</weight_inc>
        <dopamine_plasticity_time>10</dopamine_plasticity_time>
        <minweight>-0.7</minweight>
        <maxweight>0.864249</maxweight>
        <three_factor_plasticity></three_factor_plasticity>
        <maxTSSISI>10</maxTSSISI>
          </props>
        </Section>
        <Section name="WTA">
          <props>
        <n>150</n>
        <Structure type="O" dimension="10"></Structure>
        <chartime>1</chartime>
          </props>
        </Section>
        <Section name="REWGATE">
          <props>
        <n>150</n>
        <Structure type="O" dimension="10"></Structure>
        <chartime>1</chartime>
          </props>
        </Section>
        <Section name="OUT">
          <props>
        <n>10</n>
        <chartime>1</chartime>
          </props>
        </Section>
```

```xml
    <Section name="BIASGATE">
      <props>
<n>10</n>
<chartime>1</chartime>
      </props>
    </Section>
    <Link from="R" to="L" type="plastic">
      <IniResource type="uni">
<min>1.267</min>
<max>1.267</max>
      </IniResource>
        <probability>1</probability>
      <maxnpre>1000</maxnpre>
    </Link>
    <Link from="L" to="WTA" policy="aligned">
      <weight>9</weight>
    </Link>
    <Link from="WTA" to="WTA" policy="all-to-all-sections" type="gating">
      <weight>-10</weight>
    </Link>
    <Link from="WTA" to="REWGATE" policy="aligned" type="gating">
      <weight>1</weight>
    </Link>
    <Link from="REWGATE" to="L" policy="aligned" type="reward">
      <weight>0.042</weight>
    </Link>
    <Link from="WTA" to="OUT" policy="aligned">
      <weight>10</weight>
    </Link>
    <Link from="OUT" to="BIASGATE" policy="aligned" type="gating">
      <weight>-20</weight>
    </Link>
    <Link from="Target" to="REWGATE" policy="aligned">
      <weight>10</weight>
    </Link>
    <Link from="Target" to="BIASGATE" policy="aligned">
      <weight>10</weight>
      <Delay type="uni">
<min>10</min>
<max>10</max>
      </Delay>
    </Link>
    <Link from="Target" to="BIASGATE" policy="exclusive">
      <weight>-30</weight>
    </Link>
    <Link from="BIASGATE" to="L" policy="aligned">
      <weight>3</weight>
    </Link>
  </Sections>
</NETWORK>
  <Readout lib="StateClassifier">
    <output>OUT</output>
  </Readout>
</SNN>
```

# Appendix B. The Python Script Used to Compare CoLaNET Accuracy with Traditional ML Algorithms

```python
# -*- coding: utf-8 -*-
"""
Created on Fri Apr 26 16:07:53 2024

@author: Mike
"""

import numpy as np
from sklearn.ensemble import RandomForestClassifier
from sklearn.neural_network import MLPClassifier

image_file = "MNIST.bin"
target_file = "MNIST.target"
nlearningpictures = 60000
ninputs = 28 * 28
nclasses = 10

def read_data(filX, strY, ncols, nrows):
    X = np.zeros((nrows, ncols))
    Y = np.zeros(nrows)
    for i in range(nrows):
        s = filX.read(ncols)
        if len(s) < ncols:
            print("Too few records in ", image_file)
            exit(-1)
        X[i] = np.frombuffer(s, dtype=np.uint8)
        s = strY[i]
        Y[i] = int(s)
    return X, Y

with open(image_file, "rb") as filimage, open(target_file) as filtargets:
    targets = filtargets.readlines()
    Xtrain, Ytrain = read_data(filimage, targets, ninputs, nlearningpictures)
    Xtest, Ytest = read_data(filimage, targets[nlearningpictures:], ninputs, len(targets) -
     nlearningpictures)

print("Data read...")

clf = RandomForestClassifier(n_estimators=150, max_leaf_nodes=769)
clf.fit(Xtrain, Ytrain)
print("Baseline accuracy (RF): ", clf.score(Xtest, Ytest))
pred = clf.predict(Xtest)
real = np.zeros(nclasses)
predicted = np.zeros(nclasses)
correcty_predicted = np.zeros(nclasses)
for i in range(len(Ytest)):
    p = int(pred[i])
    real[int(Ytest[i])] += 1
    predicted[p] += 1
    if p == Ytest[i]:
        correcty_predicted[p] += 1
for cla in range(nclasses):
    rPrecision = correcty_predicted[cla] / predicted[cla] if predicted[cla] > 0 else  0.
    rRecall = correcty_predicted[cla] / real[cla] if real[cla] > 0 else 0.
    print("class ", cla, "precision = ", rPrecision, " recall = ", rRecall, " F = ", 2 / (1 /
     rPrecision + 1 / rRecall) if rPrecision * rRecall > 0 else 0.)

clf = MLPClassifier(hidden_layer_sizes=(150,), random_state=1, max_iter=300)
clf.fit(Xtrain, Ytrain)
print("Baseline accuracy (MLP): ", clf.score(Xtest, Ytest))
pred = clf.predict(Xtest)
real = np.zeros(nclasses)
predicted = np.zeros(nclasses)
correcty_predicted = np.zeros(nclasses)
for i in range(len(Ytest)):
    p = int(pred[i])
    real[int(Ytest[i])] += 1
    if p > 0:
        predicted[p] += 1
        if p == Ytest[i]:
            correcty_predicted[p] += 1
```

```
for cla in range(nclasses):
    rPrecision = correcty_predicted[cla] / predicted[cla] if predicted[cla] > 0 else  0.
    rRecall = correcty_predicted[cla] / real[cla] if real[cla] > 0 else 0.
    print("class ", cla, "precision = ", rPrecision, " recall = ", rRecall, " F = ", 2 / (1 /
    rPrecision + 1 / rRecall) if rPrecision * rRecall > 0 else 0.)
```